\documentclass[11pt]{article}
\usepackage{fullpage}
\usepackage{graphicx}
\usepackage{color}
\usepackage[table]{xcolor}
\usepackage{rotating}
\usepackage{tabularx}
\usepackage{pdflscape}
\usepackage{amsmath}
\usepackage{amsfonts}
\usepackage{algorithm}
\usepackage{algpseudocode}
\usepackage{bbm}

\begin{document}

\title{A Note on Information-Directed Sampling and Thompson Sampling}

\author{Li Zhou \\ lizhou@cs.cmu.edu \\ Carnegie Mellon University}

\date{March 13, 2015}
\maketitle
\begin{abstract}
  This note introduce three Bayesian style Multi-armed bandit algorithms:
  Information-directed sampling, Thompson Sampling and Generalized
  Thompson Sampling. The goal is to give an intuitive explanation for these three
  algorithms and their regret bounds, and provide some derivations that are
  omitted in the original papers.
\end{abstract}

\section{Introduction}
A multi-armed bandit problem \cite{bubeck2012regret} is one of the sequential
decision making problem. At each time the learner selects an action based on its
current knowledge and arm-selection policy, and then receives reward of the
action selected. Since the rewards of actions that are not selected are unknown,
the learner needs to balance between exploit its current knowledge to select a
best arm and explore potential best arms. In this note we describe three
Bayesian style Multi-armed bandit algorithms: Information-Directed
Sampling\cite{russo2014learning}, Thompson Sampling\cite{chapelle2011empirical}
and Generalized Thompson Sampling\cite{li2013generalized}. Each of these three
algorithms maintains a posterior distribution indicating the probability of each
arm/policy being optimal. However they have different rules to update this
posterior distribution based on observed rewards. 
\section{Information-Directed Sampling}
\subsection{Problem Formulation}
Information-Directed Sampling (IDS) \cite{russo2014learning} consider a Bayesian
formulation of
Multi-armed bandit problem. In this setting there is a set of actions (arms)
$\mathcal{A}$, and at time $t \in [1,T]$ the decision-maker chooses
an action $a_t$. Action $a_t$ then draws a reward $r_{a, t}$ from a reward
distribution $p_a$\footnote{In the original paper they assume that the arms will
first draw an outcome from an outcome distribution, then here is a fixed and
known function that maps outcomes to rewards. However here for the sake of
simplicity, we assume the outcome is equal to the reward.}. We assume that all
rewards are i.i.d distributed and the reward distribution is stationary with
respect to time $t \in [1, T]$.

To formulate Multi-armed bandit in a Bayesian way, We denote
$a^* = \arg\max_{a\in\mathcal{A}} \mathbb{E}_{r_a \sim p_a}[r_a]$, which means
$a^*$ is the arm with highest expected reward with respect to distribution
$p_a$, where $a\in\mathcal{A}$. We also denote $r_{a^*}$ the reward drawn from
$p_{a^*}$. The decision-maker do not know the real reward distribution $p_a$, so
it has its own estimate about these distributions at time step $t$, which we
denote as $\hat{p}_{a, t}$. Because of this uncertainly, 
for each action $a$ at time $t$, the decision-maker has a believe on whether
this action has the highest expected reward. We denote this believe by
$\alpha_t(a) = P(a^* = a|\mathcal{F}_{t-1})$, where $\mathcal{F}_{t-1}$ is the
history of past observations including the actions selected and the
corresponding rewards. The decision-maker will update this posterior
distribution at each time step based on $\mathcal{F}_{t-1}$.

Instead of sampling actions directly based on posterior distribution
$\alpha_t$, IDS sample actions based on a distribution $\pi$. $\pi$ is also a
distribution over all actions and is constructed based on the posterior
distribution $\alpha_t$. We are interested in the following expected regret
\begin{align}
\label{regret_definition}
  \mathbb{E}[\text{Regret}(T)] = \underset{r_{a^*} \sim p_{a^*}}{\mathbb{E}}
  \sum_{t=1}^T r_{a^*} - \underset{\begin{subarray}{c}
  a \sim \pi \\
  r_{a, t} \sim p_{a}
  \end{subarray}}{\mathbb{E}} \sum_{t=1}^T r_{a, t}
\end{align}
\subsection{Algorithm}
In multi-armed bandit problem, we want to balance between exploitation and
exploration. IDS handle this trade-off by defining immediate regret
$\bigtriangleup_t(a)$ and information gain $g_t(a)$ of action $a$ at time
$t$.
\subsubsection{Immediate Regret}
The immediate regret $\bigtriangleup_t(a)$ is defined as
\begin{align}
\bigtriangleup_t(a) = \underset{\begin{subarray}{c}
a^* \sim \alpha_t \\
r_{a^*,t} \sim \hat{p}_{a^*, t}
\end{subarray}
}{\mathbb{E}} [r_{a^*, t} | \mathcal{F}_{t-1}] - \underset{r_{a, t} \sim
  \hat{p}_{a, t}}{\mathbb{E}} [r_{a, t} | \mathcal{F}_{t-1}]
\end{align}
The idea behind this is that: the regret is defined by formula
(\ref{regret_definition}), however the decision-maker does not know the true
$p_{a^*}$ and $p_a$ for $a\in\mathcal{A}$, so it uses $\hat{p}_{a^*}$ and
$\hat{p}_a$ instead to estimate the regret at time step t. 
Note that
\begin{align}
P(r_{a^*, t} = r) = P(r_{a, t} = r | a^* = a) 
\end{align}
So
\begin{align}
\mathbb{E}[r_{a^*, t} | \mathcal{F}_{t-1}] = E[r_{a, t} | r_{b, t} \le r_{a, t}
  \text{\ }
  \forall b, \mathcal{F}_{t-1}]
\end{align}
We will show how to
calculate each of these terms in section \ref{beta_bern_exp}.
\subsubsection{Information Gain}
Instead of doing pure exploitation using immediate regret, one would want to do
some exploration to seek potential best arms. To do this, IDS defined a term:
information gain, denoted as $g_t(a)$. The idea is that: we already have a
posterior distribution over $a^*$, we hope that after we pull one of the arms,
the entropy of this distribution decreases, so that we gain a certain amount of
information about which arm has the highest expected reward. Let $a^*_t \sim
\alpha_t$ and $a^*_{t+1} \sim \alpha_{t+1}$, and let $H(a^*_t)$
denote the entropy of $a^*_t$, then $g_t(a)$ is defined
as
\begin{align}
g_t(a) = \mathbb{E}[H(a^*_t) - H(a^*_{t+1}) | \mathcal{F}_{t-1}, a_t = a]
\end{align}
The expectation is with respect to the random reward of arm $a$. To calculate
this, one can sample reward from $\hat{p_a}$ and then calculate the expectation
above. However in the original paper they used the following way.

\begin{figure}[h]
\caption{Example of $\hat{p}_{a, t}(\cdot| a')$ with 3 arms}
\label{img:pdf}
\includegraphics[scale=0.5]{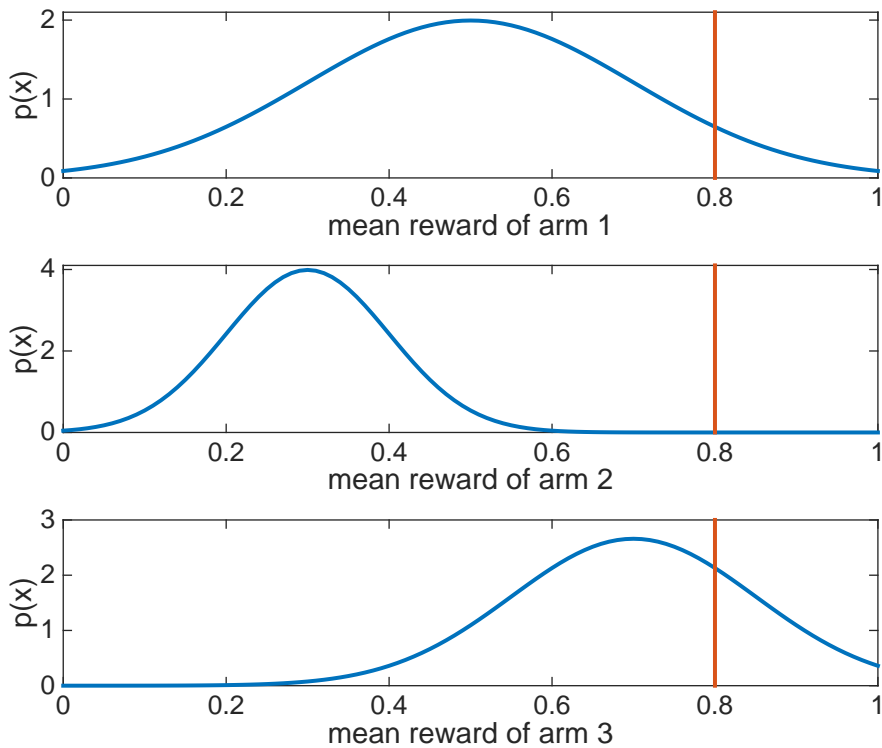}
\centering
\end{figure}

From the property of mutual information we have:
\begin{align}
H(X) - H(X|Y) = I(X, Y)
\end{align}
and since $\mathbb{E}[H(a^*_{t+1})|\mathcal{F}_{t-1}, a_t = a] =
H(a^*_{t}| r_{a,t})$, So
\begin{align}
g_t(a) = I(a^*_t, r_{a, t})
\end{align}
Also from the property of mutual information we have:
\begin{align}
I(X, Y) = \mathbb{E} D_{KL}(P(Y|X)||P(Y))
\end{align}
Since we do not have the true distribution of $r_{a, t}$, we use the posterior
distribution $\hat{p}_{a, t}$, and we have:
\begin{align}
g_t(a) = \underset{a' \sim \alpha_t}{\mathbb{E}} D_{KL}(\hat{p}_{a, t}(\cdot| a')||\hat{p}_{a, t})
\end{align}
In the equation above, $\hat{p}_{a, t}$ is just the reward posterior
distribution of arm $a$ at time $t$, and $\hat{p}_{a, t}(\cdot| a')$ is the
reward posterior distribution conditioned on that $a'$ is the arm that has the
highest mean reward. With this condition, the reward posterior distribution has
to shift to satisfy this constrain. For example in Figure \ref{img:pdf}, we show
3 arms with mean reward as Gaussian distribution, suppose we want to calculate
the reward posterior distribution of arm 2 and 3 conditioned on that arm 1 has the
highest mean reward. We examine one point where the mean reward of arm 1 is
0.8. Then the mean reward of arm 2 and arm 3 cannot be greater than 0.8, so the
probability mass of these two arms that is greater than 0.8 has to be cut off,
and the remaining has to be normalized.

\subsubsection{Optimization}
The goal of IDS at a single time step is to balance immediate regret
$\bigtriangleup_t(a)$ and information gain $g_t(a)$. There are many ways to do
this, and in the paper the author choose the following way:
\begin{align}
\pi_t^{IDS} = {\arg\min}_{\pi \in \mathcal{D}(\mathcal{A})} \left\{ \Psi_t(\pi) :=
  \frac{\bigtriangleup_t(\pi)^2}{g_t(\pi)}\right\}
\end{align}
Note that $\pi$ is a distribution over all arms, and assuming $g$ has at least 1
non-zero elements, then to find $\Psi_t(\pi)$ it is equal to solve the following
optimization problem:
\begin{align}
\text{minimize\ \ \ } &\Psi(\pi) := \frac{(\pi^T\bigtriangleup)^2}{\pi^Tg} \\
\text{subject to\ \ \ } &\pi^T e = 1 \\
&\pi \ge 0
\end{align}
The author stated that $\pi$ can be very sparse, with only two non-zero
elements, and then they try all possible combinations of two arms that gives the
lowest $\Psi_t(\pi)$. Given $\pi$, IDS sample an arm and pull that arm. I omit
the detail here since it's well described in the IDS paper.
\subsection{Bernoulli Bandit Experiment}
\label{beta_bern_exp}
In a K-armed Bernoulli bandit problem, there are K arms, and the reward of
the i-th arm follows a Bernoulli distribution with mean $X_i$. In a Bayesian
style learning algorithm, it is standard to model the mean reward of each arm
using the Beta distribution:
\begin{align}
X_i &\sim Beta(\beta_i^1, \beta_i^2) \\
r_i &\sim Bernoulli(X_i)
\end{align}
To calculate $\bigtriangleup_t(a)$ and $g_t(a)$, we first calculate
$\alpha_t(a)$. Let $f_i = Beta.pdf(x|\beta_i^1, \beta_i^2)$ and $F_i =
Beta.cdf(x|\beta_i^1, \beta_i^2)$ for all arm $i$, that is, $f_i$ and $F_i$ are
the PDF and CDF of the posterior distribution of $X_i$, then to calculate
$\alpha_t$:
\begin{align}
\alpha_t(a) &= P\left(\bigcap_{j\ne i}\{X_j \le X_i\} \right) \\
&= \int_0^1 f_i(x) P\left(\bigcap_{j\ne i}\{X_j \le X_i\}|X_i=x \right) dx \\
&= \int_0^1 f_i(x) \left( \prod_{j\ne i}F_j(x) \right) dx \\
&= \int_0^1 \left[ \frac{f_i(x)}{F_i(x)} \right] \bar{F}(x) dx
\end{align}
where $\bar{F}(x) = \prod_{i=1}^K F_i(x)$. To calculate this integral, we need
to sample points from $f_i$, $F_i$ and $\bar{F}_i$, and then do summation, so it
is quite time consuming.

Next we need to calculate $\hat{p}_{a, t}(\cdot|a^* = a)$, which is the same as
calculating $M_{ij} := E[X_j|X_k \le X_i \text{\ } \forall k]$
\begin{align}
M_{ij} &= E[X_j|X_k \le X_i \ \forall k] \\
&= \int_0^1 x P(X_j = x | X_k \le X_i \ \forall k) \\
&= \int_0^1 x \frac{P(X_j = x, X_k \le X_i \ \forall k)}{P(X_k \le X_i \ \forall
  k)} dx 
\label{mij}
\end{align}
Suppose $i \ne j$, then
\begin{align}
\text{(\ref{mij})} &= \frac{1}{\alpha_t(i)} \int_0^1 x P(X_k \le X_i \ \forall k \ne j, X_j=x, X_i\ge
  x)dx \\
&= \frac{1}{\alpha_t(i)} \int_0^1 x P(X_j = x)P(X_k \le X_i \ \forall k\ne i
  \text{ or } j,
  X_i \ge x) dx \\
&= \frac{1}{\alpha_t(i)} \int_0^1 x P(X_j = x)\int_x^1 P(X_k \le y \ \forall k\ne i
  \text{ or } j)P(X_i = y) dy dx \\
&= \frac{1}{\alpha_t(i)} \int_0^1 x P(X_j=x) \int_x^1\left(
  \frac{f_i(y)\bar{F}(y)}{F_i(y) F_j(y)} \right) dydx \\
&= \frac{1}{\alpha_t(i)} \int_0^1 \left( \frac{f_i(y) \bar{F}(y)}{F_i(y)F_j(y)}
  \right) \int_0^y xf_j(x)dx dy  \\
&= \frac{1}{\alpha_t(i)} \int_0^1 \left( \frac{f_i(y) \bar{F}(y)}{F_i(y)F_j(y)}
  \right) Q_j(y) dy
\end{align}
Where $Q_j(y) = \int_0^y xf_j(x)dx$. To calculate $Q_j(y)$ we also need to do
sampling and then summation.
Suppose $i = j$, then
\begin{align}
\text{(\ref{mij})} &= \frac{1}{\alpha_t(t)} \int_0^1 x P(X_i = x, X_k \le x \
                     \forall k\ne i) \\
&= \frac{1}{\alpha_t(i)} \int_0^1 x f_i(x) \prod_{j\ne i} F_j(x) dx \\
&= \frac{1}{\alpha_t(i)} \int_0^1 \frac{xf_i(x)}{F_i(x)}\bar{F}(x) dx
\end{align}
Now that we have $\alpha_t(a)$ and $M_{ij} = E[X_j|X_k \le X_i \ \forall k]$, we
can calculate $\bigtriangleup_t(a)$ and $g_t(a)$.
\begin{align}
\rho^* &= \sum_{i=1}^K \alpha_t(i) M_{ii} \\
\bigtriangleup_t(i) &= \rho^* - \frac{\beta_i^1}{\beta_i^1+\beta_i^2} \\
g_i &= \sum_{i=1}^K \alpha_j KL\left( M_{ji} || \frac{\beta_i^1}{\beta_i^1+\beta_i^2} \right)
\end{align}
Where $KL(p_1||p_2)$ is defined as $KL(p_1||p_2) = p_1\log(\frac{p_1}{p_2}) +
(1-p_1)\log(\frac{1-p_1}{1-p_2})$ since $\hat{p}_{a, t}$ follows Bernoulli
distribution.

At each time step, we can calculate $\bigtriangleup_t(a)$ and $g_t(a)$ by the
above procedure and then solve the optimization problem to get $\pi$, and sample
an arm based on $\pi$.
\subsection{Regret Bound}
Here we prove a general regret bound, for specific regret bound, we can refer to
the IDS paper.
For a fixed deterministic $\lambda \in \mathbb{R}$ and a policy $\pi$ such at
$\Psi_t(\pi_t) \le \lambda$, we have
\begin{align}
\mathbb{E}[\text{Regret}(T, \pi)] \le \sqrt{\lambda H(\alpha_1)T}
\end{align}
Prove:
\begin{align}
\mathbb{E}\sum_{t=1}^T g_t(\pi_t) &= \mathbb{E}\sum_{t=1}^T
                                    E[H(\alpha_t)-H(\alpha_{t+1})|\mathcal{F}_{t-1}]\\
&= \mathbb{E}\sum_{t=1}^T (H(\alpha_t) - H(\alpha_{t+1})) \\
&= H(\alpha_1) - \mathbb{E}H(\alpha_{T+1}) \\
&\le H(\alpha_1)
\end{align}
By definition, $\Psi_t(\pi) \le \lambda$, so $\bigtriangleup_t(\pi) \le
\sqrt{\lambda g_t(\pi)}$, so
\begin{align}
\mathbb{E}(\text{Regret}(T, \pi)) &= \mathbb{E}\sum_{t=1}^T
                                    \bigtriangleup_t(\pi) \\
&\le \sqrt{\lambda} \mathbb{E}\sum_{t=1}^T \sqrt{g_t(\pi)} \\
&\le \sqrt{\lambda T} \sqrt{\mathbb{E}\sum_{t=1}^T g_t(\pi)} \text{\ \
  Caushy-Schwardsz inequality}\\
&\le \sqrt{\lambda H(\alpha_1)T}
\end{align}
In the paper, the author proved that $\Psi_t^* \le |\mathcal{A}/2|$, so
$\mathbb{E}(\text{Regret}(T, \pi^{IDS})) \le \sqrt{\frac{1}{2}|\mathcal{A}|H(\alpha_1)T} $
\subsection{Potential Problems}
IDS showed a strong empirical results, however there are several potential
problems. I think the main problem is that the algorithm is very time consuming
as I run it, the reason is that it has 3 integral to calculate so we have to
evaluate each integrand at a discrete grid of points. Another problem is that
the paper didn't mention why they choose such format of $\Psi$ as the trade-off
between $\bigtriangleup_t$ and $g_t$, since there are many ways to make this
trade-off. Also it would be nice to see some generalization to contextual
bandit. 

\section{Thompson Sampling}
\subsection{Problem Formulation}
Thompson sampling (TS) \cite{chapelle2011empirical, agrawal2013further} is also
a Bayesian style bandit algorithm, it can apply to both contextual bandit and
standard Multi-armed bandit problems. Here we talk about the non-contextual
version. Again, we assume there is an action set $\mathcal{A}$, and at time step
$t$ Thompson sampling select action $a$ and get reward $r_{a, t}$.
We also assume the reward of each arm $r_{a}$ follows some parametric
distribution $p_{a} = P(r|a, \theta_a)$ with mean $\mu_a$, where $\theta_a$ is
the parameter. Define past observations
$\mathcal{D}$ consists of arms pulled and rewards observed. At the beginning,
Thompson sampling assumes a prior distribution on parameters $\theta_a$, and
then after each time step, it will update the posterior distribution $P(\theta_a|\mathcal{D})$ based on past
observations. Similar to IDS, the goal is to minimize the
regret:
\begin{align}
\mathbb{E}[\text{Regret}(T)] = \underset{r_{a^*} \sim p_{a^*}}{\mathbb{E}} \sum_{t=1}^T r_{a^*} - \underset{r_{a, t} \sim p_a}{\mathbb{E}}\sum_{t=1}^T r_{a, t}
\end{align}
where $a^*$ is the arm with the highest expected reward, and $a$ is the arm
selected by Thompson sampling.
\subsection{Algorithm}
Similar to IDS, Thompson sampling randomly select an action $a$ according to its
probability of being optimal. So action $a$ is chosen with probability
\begin{align}
\int \mathbb{I} \left[ E(r|a, \theta) = \max_{a'} E(r|a', \theta) \right]
  P(\theta|D) d\theta
\end{align}
Which is essential the same as the $\alpha_t$ in IDS. However
calculating $\alpha_t$ is time consuming, and since in Thompson sampling, we do not
need to use $\alpha_t$ explicitly, and we only need samples from $\alpha_t$, so it
suffices to draw a random parameter $\theta$ from posterior
distribution. Algorithm \ref{tsbern} describes the procedure of Thompson sampling
with Bernoulli bandit problem.
\begin{algorithm}
\caption{Thompson sampling with Bernoulli multi-armed bandit}
\label{tsbern}
\begin{algorithmic}
\Require $\alpha$, $\beta$: prior parameter of a Beta distribution
\State For each arm $i=1, ..., K$ set $S_i = 0$, $F_i = 0$
\For{$t=1, ..., T$}
\For{arm $i=1, ..., K$}
\State Draw $\theta_i$ from $Beta(\alpha+S_i, \beta+F_i)$
\EndFor
\State Play arm $a = \arg\max_i \theta_i$, and observe reward $r_t$
\If{$r_t=1$} 
$S_a=S_a+1$
\Else $\ F_a = F_a+1$
\EndIf 
\EndFor
\end{algorithmic}
\end{algorithm}
\subsection{Regret}
Although Thompson sampling is a very old algorithm, proposed by
\cite{thompson1933likelihood}, but the theoretical analysis is done very
recently. We follow \cite{agrawal2013further} and hope to give a intuitive
explanation of the regret. 
Let $\mu^* = \max_i \mu_i$ and $\bigtriangleup_i = \mu^* - \mu_i$, where $i \in
\mathcal{A}$, and let $k_i(t)$ denote the number of times arm $i$ has been
played up to step $t-1$. Then the expected total regret in time $T+1$ can be written as
\begin{align}
\mathbb{E}[\text{Regret}(T)] = \sum_i \bigtriangleup_i \mathbb{E}(k_i(T+1))
\end{align}
Hence to bound the expected regret, we need to bound $\mathbb{E}(k_i(T+1))$ for
all $i \in \mathcal{A}$.

To bound $k_i(T+1)$ we need the following settings \cite{agrawal2013further}:
Define $F_{n,p}^B(\cdot)$ the $cdf$ and $f_{n, p}^B(\cdot)$ the $pdf$ of the
binomial distribution with parameters $n, p$. Define $F_{\alpha,
  \beta}^{\text{beta}}(\cdot)$ the $cdf$ of beta distribution with parameters
$\alpha, \beta$. Let $i(t)$ denote the arm played at time $t$, $k_i(t)$ denotes
the number of plays of arm $i$ until time $t-1$, $S_i(t)$ denote the number
of successes among the plays of arm $i$ until $t-1$ for the Bernoulli bandit
case, $\hat{\mu}(i)$ denote the empirical mean and $\theta_i(t)$ denote the
sample mean reward of arm $i$ at time $t$.
We assume the first arm is the unique optimal arm, i.e\., $\mu^* = \mu_1$.
For each arm $i$, we will choose two thresholds $x_i$ and $y_i$ such that $\mu_i
< x_i < y_i < \mu_1$. With different choices of $x_i$ and $y_i$, we can get
problem dependent and problem independent bound respectively. We also define
$E^\mu_i(t)$ as the event that $\hat{\mu}_i(t) \le x_i$ and $E^\theta_i(t)$ as
the event that $\theta_i(t) \le y_i$. Finally, define $\mathcal{F}_{t-1} = \{i(w),
  r_{i(w)}(w), w=1,..., t-1\}$ and $p_{i, t} = P(\theta_1(t) > y_i |
  \mathcal{F}_{t-1})$. $p_{i, t}$ indicates what is the probability of the
  sample reward of arm $1$ is greater than $y_i$ at time $t$.

We can decompose $\mathbb{E}(k_i(T+1))$ into
\begin{align}
\mathbb{E}[k_i(T+1)] &= \sum_{t=1}^T P(i(t) = i) \\
& = \sum_{t=1}^T P(i(t) = i, E_i^{\mu}(t), E_i^{\theta}(t))  \label{ekt_decomp1} \\
& \quad+ \sum_{t=1}^T P(i(t)=i, E_i^{\mu}(t),
  \overline{E_i^{\theta}(t)}) \label{ekt_decomp2} \\
& \label{ekt_decomp3} \quad+ \sum_{t=1}^T P(i(t)=i, \overline{E_i^{\mu}(t)})
\end{align}
So we need to bound (\ref{ekt_decomp1}), (\ref{ekt_decomp2}) and (\ref{ekt_decomp3}) respectively.
To bound (\ref{ekt_decomp1}), \cite{agrawal2013further} proved that
\begin{align}
P(i(t)=i, E_i^\mu(t), E_i^\theta(t) | \mathcal{F}_{t-1}) \le \frac{(1-p_{i,
  t})}{p_{i, t}} P(i(t)=1, E_i^\mu(t), E_i^\theta(t) | \mathcal{F}_{t-1})
\end{align}
and so
\begin{align}
  \sum_{t=1}^T P(i(t)=i, E_i^\mu(t), E_i^\theta(t)) &= \sum_{t=1}^T
  \mathbb{E}P(i(t)=i, E_i^\mu(t), E_i^\theta(t) | \mathcal{F}_{t-1}) \\
  &\le \sum_{t=1}^T \mathbb{E} \left[ \frac{(1-p_{i,t})}{p_{i, t}} P(i(t)=1,
  E_i^\mu(t), E_i^\theta(t) | \mathcal{F}_{t-1}) \right] \\
  & \le \sum_{k=0}^{T-1}\mathbb{E}\left[ \frac{1}{p_{i, \tau_k+1}}-1\right]
\label{ekt_decomp1:1}
\end{align}
where $\tau_k$ denotes the time step at which arm 1 is played for the $k^{th}$
time. (\ref{ekt_decomp1:1}) only involves $p_{i, \tau_k+1}$ because the
posterior distribution of the parameters of arm 1 only changes when arm 1 gets
pulled. Now we need to bound (\ref{ekt_decomp1:1}). Let $k_1(t)=j$ and
$S_1(t)=s$, from the fact that $F_{\alpha, \beta}^{beta}(y) = 1 -
F_{\alpha+\beta-1,y}^B(\alpha-1)$ we have $p_{i, t} =
P(\theta_1(t)>y_i)=F_{j+1}^B(s)$, and since
\begin{align}
S_1(t) \sim \text{Binomial}(k_1(t), \mu_1) \\
\theta_1(t) \sim \text{Beta}(S_1(t), k_1(t)-S_1(t))
\end{align}
so each possible value $S_1(t) = s$ corresponding to a value of
$p_{i, \tau_j+1} =F_{j+1, y}^B(s)$ with probability $f_{j, \mu_1}^{B}(s)$, so 
\begin{align}
\mathbb{E}\left[ \frac{1}{p_{i, \tau_k+1}}-1\right] = \sum_{s=0}^j \frac{f_{j,
  \mu_1}^B(s)}{F_{j+1, y}^B(s)}
\label{ekt_decomp1:2}
\end{align}
So we have reduced the problem of bounding (\ref{ekt_decomp1:1}) to the problem
of bounding a summation of a series of random variables involving binomial
distribution. \cite{agrawal2013further} provide details about how to bound
(\ref{ekt_decomp1:2}), which is quite complicated.

Now we bound (\ref{ekt_decomp3}). Let $\tau_k$ denote the time at which $k^{th}$
trial of arm $i$ happens, and $\tau_0 = 0$. We have
\begin{align}
\sum_{t=1}^T P(i(t) = i, \overline{E_i^\mu(t)}) &\le \mathbb{E}\left[
  \sum_{k=0}^{T-1} \sum_{t=\tau_k+1}^{\tau_{k+1}} I(i(t)=i)
  I(\overline{E_i^\mu(t)}) \right] \label{ekt_decomp3:1}
\end{align}
Since $E_i^\mu(t)$ doesn't change unless arm $i$ is pulled, and
$\sum_{t=\tau_k+1}^{\tau_{k+1}} I(i(t)=i) = 1$, so
(\ref{ekt_decomp3:1}) is equal to
\begin{align}
(\ref{ekt_decomp3:1}) &= \mathbb{E}\left[ \sum_{k=0}^{T-1}
                      I(\overline{E_i^\mu(\tau_k+1)})
                      \sum_{t=\tau_k+1}^{\tau_{k+1}} I(i(t)=i)  \right] \\
&= \mathbb{E}\left[ \sum_{k=0}^{T-1}I(\overline{E_i^\mu(\tau_k+1)}) \right] \\
&\le 1 + \mathbb{E} \left[ \sum_{k=1}^{T-1}I(\overline{E_i^\mu(\tau_k+1)})
  \right] \\
&\le 1 + \sum_{k=1}^{T-1} \exp(-kd(x_i, \mu_i)) \\
&\le 1 + \frac{1}{d(x_i, \mu_i)}
\end{align}
Where the second last inequality is from Chernoff bound and $d(x, y) =
x\ln\frac{x}{y} + (1-x)\ln\frac{(1-x)}{(1-y)}$.

Similarly, \cite{agrawal2013further} bound 
\begin{align}
\sum_{t=1}^T P(i(t)=i, \overline{E_i^\theta(t)}, E_i^\mu(t)) \le L_i(T) + 1
\end{align}
where $L_i(T) = \frac{\ln T}{d(x_i, y_i)}$
Together with this three bounds and a choice of $x_i$ and $y_i$ for all $i\in
\mathcal{A}$, we can get a problem independent bound $O(\sqrt{NT\ln N})$.

\section{Generalized Thompson Sampling}
\subsection{Problem Formulation}
Generalized Thompson Sampling\cite{li2013generalized} is a contextual bandit
problem, it is similar to expert-learning framework, and include Thompson
Sampling as a special case. Let $\mathcal{X}$ and $\mathcal{A}$ be the set of
context and arms, and let $K=|\mathcal{A}|$. At time step $t\in{1, ..., T}$, the
decision-maker observes the context $x_t \in \mathcal{X}$ and selects an arm
$a_t \in \mathcal{A}$. Then it receives reward $r_t \in \{0, 1\}$, with
expectation $\mu(x_t, a_t)$. In \cite{li2013generalized} the reward is binary,
but it is easy to generalize to continuous space. Different from classic
Thompson Sampling algorithm, Generalized Thompson Sampling allows the
decision-maker to have access to a set of experts $\mathcal{E}=\{\mathcal{E}_1,
  \mathcal{E}_2, ..., \mathcal{E}_n\}$, each $\mathcal{E}$ makes predicts about
the average reward $\mu(x_t, a_t)$. Let $f_i$ be the associated prediction
function of expert $\mathcal{E}_i$, the arm-selection policy is
$\mathcal{E}_i(x) = \max_{a\in\mathcal{A}}f_i(x, a)$. Each expert could be a
generalized linear model or other prediction model. The regret is defined as
\begin{align}
\mathbb{E}\left[ \text{Regret}(T) \right] = \max_{1\le i\le N} \sum_{t=1}^N
  \mu(x_t, \mathcal{E}_i(x_i)) - \mathbb{E}\left[ \sum_{t=1}^T \mu(x_t, a_t) \right]
\end{align}
That is, we are competing with the best expert.
\subsection{Algorithm}
Generalized Thompson Sampling is described in Algorithm \ref{gtsalgo}. We can
see that it updates the weight $w_{i, t+1}$ by $w_{i, t+1} \propto w_{i, t}
\exp(-\eta \ell(f_i(x_t, a_t), r_t))$, where $\ell$ is the loss function. The term
`Generalized' in `Generalized Thompson Sampling' means that we can use different
types of loss functions when updating $w_i$. \cite{li2013generalized} described
two loss functions: logarithmic loss and square loss. Logarithmic loss is defined as
$\ell(\hat{r}, r) =
\mathbbm{1}(r=1)\ln1/\hat{r}+\mathbbm{1}(r=0)\ln(1/(1-\hat{r}))$, and square loss is
defined as $\ell(\hat{r}, r) = (\hat{r} - r)^2$. In next section, we will show that
if the loss function is logarithmic loss, then Generalized Thompson Sampling
takes the form of Thompson Sampling.
\begin{algorithm}
\caption{Generalized Thompson Sampling}
\label{gtsalgo}
\begin{algorithmic}
\Require $\eta > 0$, $\gamma > 0$, ${\mathcal{E}_1, ..., \mathcal{E}_N}$, prior $\boldsymbol{p}$
\State For each expert $i=1, ..., N$ set $\boldsymbol{w}_1 = \boldsymbol{p}$,
$W_1 = ||\boldsymbol{w}_1||_1$
\For{$t=1, ..., T$}
\State Receive context $x_t \in \mathbb{X}$
\For{arm $a=1, ..., K$}
\State $P(a) = (1-\gamma) \sum_{i=1}^N \frac{w_{i,
    t}\mathbbm{1}(\mathcal{E}_i(x_t)=a)}{W_t} + \frac{\gamma}{K}$
\EndFor
\State Select arm $a_t$ based on $P(a)$, observe reward $r_t$, update weights:
\State $\forall i: $ $w_{i, t+1} = w_{i, t} \exp(-\eta \ell(f_i(x_t, a_t), r_t)); W_{t+1} =
\sum_i{w_{i, t+1}}$
\EndFor
\end{algorithmic}
\end{algorithm}
\subsection{Connection with Expert-Learning and Thompson Sampling}
Generalized Thompson sampling has the format of expert exponential weighting,
however it also fits Thompson sampling framework, there are two ways to see
this, and in both ways we need to assume the loss is log loss, that is
if an expert $f$ predicts that the probability of $r = 1$ is $p_1$ and the
probability of $r=0$ is $1-p_1$, then the log loss of expert $f$ is
$\ln\frac{1}{p_1}$ when reward is $1$, and is $\ln\frac{1}{1-p1}$ when reward is
$0$.

The first way to see this: we can think of Generalized Thompson Sampling as
maintaining a posterior distribution of the weight of each expert,
denoted as $w_t$. This posterior distribution may be interpreted as the
posterior probability that $f_i$ is the reward-maximizing expert. The update
rule, for one step, is
\begin{align}
w_{i, t+1} &\propto w_{i, t} \exp(-\ell(f_i(x_t, a_t), r_t)) \\
&\propto w_{i, t}\exp(-\ln(\frac{1}{p(r_t|x_t, a_t)})) \\
&\propto w_{i, t}p_i(r_t|x_t, a_t)
\end{align}
Let $f^*=f_i$ be the event that $f_i$ is the reward-maximizing expert. From
Bayesian rule we have, for one step 
\begin{align}
P(f^*_{t+1} = f_i|x_t, a_t, r_t) &\propto P(r=r_t|f^* = f_i, x_t, a_t) P(f^* = f_i) \\
&\propto w_{i,t}p_i(r_t|x_t, a_t)
\end{align}
We can see that the update rule and Bayesian rule take the same format.
Finally, the posterior distribution on $f_t^*$ is
\begin{align}
P(f_t^*=f_i) = \frac{w_{i,t}}{\sum_i w_{i, t}}
\end{align}

We can also see it from a second way. Let $y_t$, $x_t$ and $r_t$ be the selected
arm, context and reward in time $t$,
$y^t$, $x^t$, $r^t$ be the selected arms, contexts and rewards in time $1, ...,
t$ respectively, then from Bayesian rule we have
\begin{align}
  p(y_t|y^{t-1}, r^{t-1}, x^t) &= \frac{p(y^{t-1}, y_t, r^{t-1},
                                 x^t)}{p(y^{t-1}, r^{t-1}, x^t)} \\
  &= \frac{p(y^t|r^{t-1}, x^t)}{p(y^{t-1}|r^{t-1}, x^t)}
\end{align}
Assume we have a uniform mixture of the distribution defined by the experts
(Note that we are assuming uniform mixture over $y^t$ and $y^{t-1}$, not
$y_t$), then we have
\begin{align}
  \frac{p(y^t|r^{t-1}, x^t)}{p(y^{t-1}|r^{t-1}, x^t)} &= \frac{\sum_f
                                                        f(y^t|r^{t-1},
                                                        x^t)}{\sum_f f(y^{t-1} |
                                                        r^{t-1}, x^t)}
\end{align}
From update rule we have:
\begin{align}
p(y_t|y^{t-1}, r^{t-1}, x^t) &= \frac{\sum_f w_{f,t-1} f(y_t|x_t)}{W_{t-1}} \\
&= \frac{\sum_f w_{f, t-1} f(y_t|x_t)}{\sum_f{w_{f, t-1}}} \\
&= \frac{\sum_f\ w_0\ p_f(r_1|y^1, x^1)\ p_f(r_2|y^2, x^2, r^1)...\
  p_f(r_{t-1}|y^{t-1}, x^{t-1}, r^{t-2})\ f(y_t|x_t)}{\sum_f{w_0\ p_f(r_1|y^1, x^1)\
  p_f(r_2| y^2, x^2, r^1)...\ p_f(r_{t-1}|y^{t-1}, x^{t-1}, r^{t-2})}} \\
&=\frac{\sum_f f(y^t, x^{t-1}, r^{t-1}|x_t)}{\sum_f{f(y^{t-1}, x^{t-1}, r^{t-1})}} \\
&=\frac{\sum_f f(y^t|x^t, r^{t-1})}{\sum_f{f(y^{t-1}|x^{t-1}, r^{t-1})}}
\end{align}
So we can see that the update rule and Bayesian rule have the same
format. However notice that in this view we are conditioned on $x^t$ while in the
first view the posterior distribution of $w_t$ is conditioned on the $x^{t-1}$.
\subsection{Regret}
The basic idea of the derivation is that we assume a connection between the
loss function and the regret: Define immediate regret $\bigtriangleup_i(x) = \mu(x,
\mathcal{E}^*(x)) - \mu(x,\mathcal{E}_i(x))$, shifted loss of expert i
$\hat{l}_i(r|x, a) = \ell(f_i(x, a), r) - \ell(f^*(x, a), r)$, and average
shifted loss $\bar{l}=\mathbb{E}_{r_t, a_t}\left[ \sum_i w_{i,
    t}\hat{l}_i(r_t|x_t, a_t) \right]$, we assume there is a constant $k_1$, such
that $\bigtriangleup_i(x_t) \le k_1 \sqrt{\bar{l}_t}$. Also we make use of the
self-boundedness property of the loss function: $\mathbb{E}_r\left[
  \hat{l}_i(r|x, a)^2 \right] \le k_2 \mathbb{E}_r\left[ \hat{l}_i(r|x,a)
\right]$, which means the second moment is bounded by
the first moment of the shifted loss. Then we can bound the expected regret by
\begin{align}
\sqrt{4k_2(e-2)}k_1(1-\gamma)\sqrt{T \cdot \ln\frac{1}{p_1}} + \gamma T
\end{align}
Different loss has different choice of $k_1$ and $k_2$, and
\cite{li2013generalized} proved that with square loss the expected regret bound
is $O(\sqrt{\ln\frac{1}{p_1}}K^{1/3}T^{2/3})$ and with logarithmic loss the expected
regret bound is $O(\sqrt{\ln\frac{1}{p_1}}K^{2/3}T^{2/3})$.

\bibliographystyle{unsrt}
\bibliography{ref}
\end{document}